\documentclass[10pt,twocolumn,letterpaper]{article}

\usepackage{wacv}
\makeatletter
\@namedef{ver@everyshi.sty}{}
\makeatother
\usepackage{pgfplots}

\usepackage{times}
\usepackage{epsfig}
\usepackage{graphicx}
\usepackage{amsmath}
\usepackage{amssymb}

\usepackage{caption}
\usepackage{adjustbox}
\usepackage{wrapfig}
\usepackage{xcolor}
\usepackage{multirow}
\usepackage{booktabs}
\usepackage{url}
\usepackage{subfigure}
\usepackage{enumitem}
\usepackage{dblfloatfix}

\usepackage{float}
\usepackage{bbm}

\def\wacvPaperID{1207} 

\wacvfinalcopy 

\ifwacvfinal
\def\assignedStartPage{1} 
\fi

\ifwacvfinal
\usepackage[breaklinks=true,bookmarks=false]{hyperref}
\else
\usepackage[pagebackref=true,breaklinks=true,colorlinks,bookmarks=false]{hyperref}
\fi

\ifwacvfinal
\else
\fi

\begin{document}

\title{Driver Anomaly Detection: A Dataset and Contrastive Learning Approach}

\author{\parbox{16cm}{\centering
    {\large Okan K\"op\"ukl\"u \hspace{0.6cm} Jiapeng Zheng \hspace{0.6cm} Hang Xu \hspace{0.6cm} Gerhard Rigoll}\\
    {\normalsize
    \vspace{0.2cm}
    Technical University of Munich
    }}
}

\maketitle

\begin{abstract}

Distracted drivers are more likely to fail to anticipate hazards, which result in car accidents. Therefore, detecting anomalies in drivers' actions (i.e., any action deviating from normal driving) contains the utmost importance to reduce driver-related accidents. However, there are unbounded many anomalous actions that a driver can do while driving, which leads to an `open set recognition' problem. Accordingly, instead of recognizing a set of anomalous actions that are commonly defined by previous dataset providers, in this work, we propose a contrastive learning approach to learn a metric to differentiate normal driving from anomalous driving. For this task, we introduce a new video-based benchmark, the Driver Anomaly Detection (DAD) dataset, which contains normal driving videos together with a set of anomalous actions in its training set. In the test set of the DAD dataset, there are unseen anomalous actions that still need to be winnowed out from normal driving. Our method reaches  0.9673 AUC on the test set, demonstrating the effectiveness of the contrastive learning approach on the anomaly detection task. Our dataset, codes and pre-trained models are publicly available \footnote{https://github.com/okankop/Driver-Anomaly-Detection}.


\end{abstract}
\section{Introduction}

\begin{figure}[t!]
	\centering
	\includegraphics[width = 0.46\textwidth]{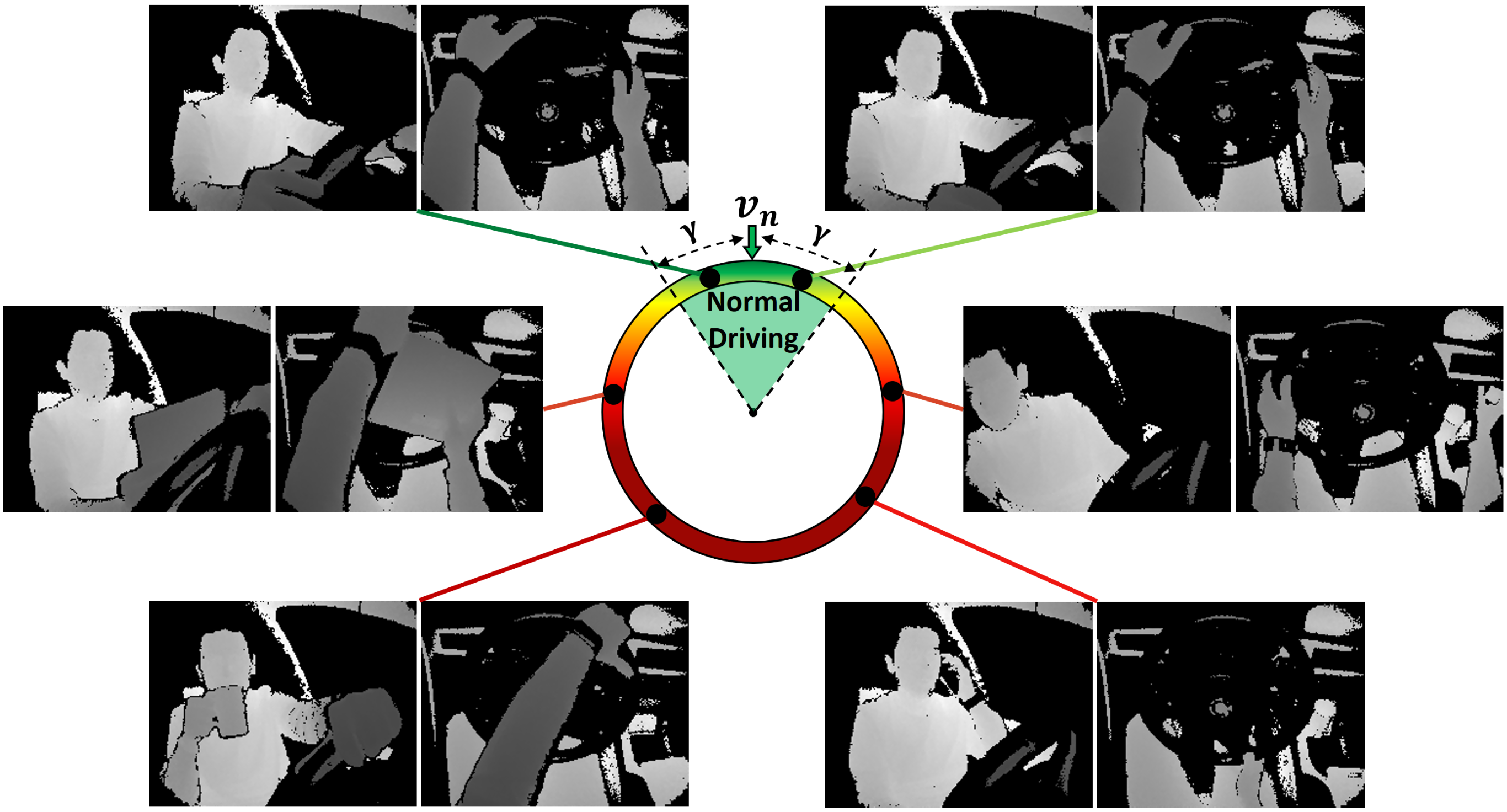}
	\caption{Using contrastive learning, normal driving template vector $\boldsymbol{v_{n}}$ is learnt during training. At test time, any clip whose embedding is deviating more than threshold $\gamma$ from normal driving template $\boldsymbol{v_{n}}$ is considered as anomalous driving. Examples are taken from new introduced Driver Anomaly Detection (DAD) dataset for front (left) and top (right) views on depth modality.}
	\vspace{-0.2cm}
	\label{fig:visual}
\end{figure}

Driving has become an indispensable part of modern life providing a high level of convenient mobility. However, this strong dependency on driving also leads to an increased number of road accidents. According to  the World Health Organization's estimates, 1.25 million people die in road accidents per year, and up to 50 million people injure. Human factors are the main contributing cause in almost 90\% of the road accidents having distraction as the main factor for around 68\% of them \cite{dingus2016driver}. Accordingly, the development of a reliable Driver Monitoring System (DMS), which can supervise a driver's performance, alertness, and driving intention, contains utmost importance to prevent human-related road accidents.

Due to the increased popularity of deep learning methods in computer vision applications, there has been several datasets to facilitate video based driver monitoring \cite{martin2019drive,ortega2020dmd,abouelnaga2017real}. However, all these datasets are partitioned into finite set of known classes, such as normal driving class and several distraction classes, with equivalent training and testing distribution. In other words, these datasets are designed for \textit{closed set recognition}, where all samples in their test set belong to one of the \textit{K} known classes that the networks are trained with. This arises a very important question: \textbf{\textit{How would the system react if an unknown class is introduced to the network?}} This obscurity is a serious problem since there might be unbounded many distracting actions that a driver can do while driving.

Different from available datasets and majority research on DMS applications, we propose an \textit{open set recognition} approach for video based driver monitoring. Since the main purpose of a DMS is to ensure that driver drives attentively and safely, which is referred as \textit{normal driving} in this work, we propose a deep contrastive learning approach to learn a metric in order to distinguish normal driving from anomalous driving. Fig. \ref{fig:visual} illustrates the proposed approach.

In order to to facilitate further research, we introduce a large scale, multi-view, multi-modal Driver Anomaly Detection (DAD) dataset. The DAD dataset contains normal driving class together with a set of anomalous driving actions in its training set. However, there are several unseen anomalous actions in the test set of DAD dataset that still need to be distinguished from normal driving. We believe that DAD dataset addresses to the true nature of driver monitoring.

Overall, the main contributions of this work can be summarized as:

\begin{itemize}
	\item We introduce DAD dataset, which is the first video based open set recognition dataset for vision based driver monitoring application. The DAD dataset is multi-view (front and top views), multi-modal (depth and infrared modalities) and large enough to train deep Convolutional Neural Network (CNN) architectures from scratch.
	\item We propose a deep contrastive learning approach to distinguish normal driving from anomalous driving. Although contrastive learning has been popular for unsupervised metric learning recently, we prove its effectiveness by achieving 0.9673 AUC in the test set of DAD dataset.
	\item We present a detailed ablation study on the DAD dataset and proposed contrastive learning approach in order give better insights about them.
\end{itemize}

\section{Related Work}

\textbf{Vision Based Driver Monitoring Datasets}. There are several hand-focused datasets such as CVRR-HANDS 3D \cite{ohn2013driver}, VIVA-Hands \cite{das2015performance} and DriverMHG \cite{kopuklu2020drivermhg}. Although these datasets aim to facilitate research on hand gesture recognition for human machine interaction, they can be used to detect hand position \cite{le2017robust}, which is highly correlated to the drivers' ability to drive. Ohn-bar \textit{et al.} introduces additional two datasets \cite{ohn2013driver,ohn2013vehicle} in order to study hand activity and pose which can be used to identify driver's state. 

Drivers' face and head information also provides very important cues to identify driver's state such as head pose, gaze directions, fatigue and emotions. There are several datasets such as \cite{agtzidis2019360,palazzi2018predicting,fang2019dada} that provide eye-tracking annotations. This information together with the interior design of the cabin help identifying where the driver is paying attention, as in DrivFace dataset \cite{diaz2016reduced}. In addition, datasets such as DriveAHead \cite{schwarz2017driveahead} and DD-Pose \cite{roth2019dd} provide head pose annotations of yaw, pitch and roll angles. 

There are also datasets that focus on the body actions of the drivers. StateFarm \cite{Kaggle} is the first image-based dataset for this purpose, which contains safe driving and 9 additional distracting classes. A similar image-based dataset AUC Distracted Driver (AUC DD) \cite{abouelnaga2017real} is proposed using a side-view camera to capture drivers' actions. However, these two datasets are image-base and lack important temporal information. A simple modification on AUC DD dataset to investigate importance of spatio-temporal information is presented in \cite{kose2019real}. Recently, Drive\&Act  dataset is introduced in \cite{martin2019drive}, which is recorded for 5 NIR cameras where subjects perform distraction-related actions for autonomous driving scenario.

None of the datasets mentioned above is designed for open set recognition scenarios \cite{6365193}, where unknown actions are performed at the test time. In this perspective, the introduced DAD dataset is the first available dataset designed for open-set-recognition.

\textbf{Contrastive Learning Approaches.} Since its initial proposition \cite{hadsell2006dimensionality}, these approaches learn representations by contrasting positive pairs against negative pairs. In \cite{wu2018unsupervised}, the full softmax distribution is approximated by the Noise Contrastive Estimation (NCE) \cite{Gutmann2010NoisecontrastiveEA}; a memory bank and the Proximal Regularization \cite{8187362} are used in order to stabilize learning process. Following works use similar approaches with several modifications. In \cite{zhuang2019local}, instances that are close to each other on the embedding space used as positive pairs in addition to the augmented version of the original images. In \cite{he2019momentum}, a dynamic dictionary with a queue and a moving-average encoder are presented. Authors in \cite{tian2019contrastive} try to bring different views of the same scene together in embedding space, while pushing views of different scenes apart. A projection head is introduced in \cite{chen2020simple}, which improves the quality of the learned representations. It has been proven that models with unsupervised pretraining achieves better than models with supervised pretraining in various tasks \cite{chen2020simple}. Moreover, performance of supervised contrastive learning is also validated in \cite{khosla2020supervised}.

\textbf{Lightweight CNN Architectures.} Since DMS applications need to be deployed in car, it is critical to have a resource efficient architecture. In recent years, several lightweight CNN architectures are proposed. SqueezeNet \cite{iandola2016squeezenet} is the first and most well-know architecture, which consists of fire modules to achieve AlexNet-level accuracy with 50x fewer parameters. MobileNet \cite{howard2017mobilenets} contains depthwise separable convolutions with a width multiplier parameter to achieve thinner or wider network. MobileNetV2 \cite{sandler2018mobilenetv2} contains inverted residuals blocks and ReLU6 activation function. ShuffleNet \cite{zhang2018shufflenet} proposes to use channel shuffle operation together with pointwise group convolution. ShuffleNetV2 \cite{ma2018shufflenet} upgrades it with several principles, which are effective in designing lightweight architectures. Networks using Neural Architecture Search (NAS) \cite{zoph2016neural}, such as NASNet\cite{zoph2018learning}, FBNet\cite{wu2019fbnet}, provide another direction for designing lightweight architectures. In this work, we have used 3D version of several resource efficient architectures, which are introduced in \cite{kopuklu2019resource}.

\section{Driver Anomaly Detection (DAD) Dataset}

\begin{figure}[t!]
    \centering
    \subfigure[Camera placements in the simulator]{
        \includegraphics[width=1.8in]{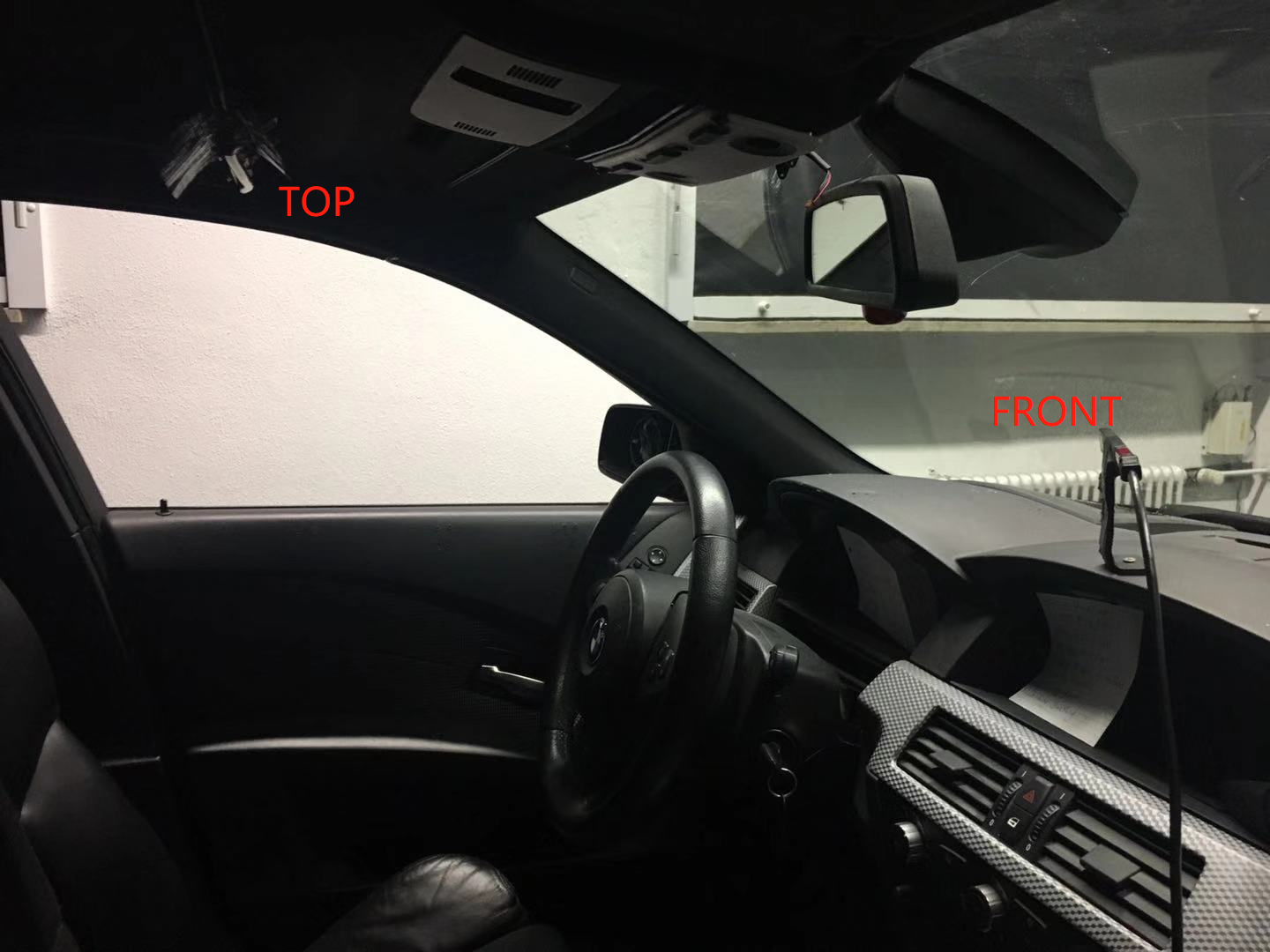}}
    \subfigure[Camera]{
        \includegraphics[width=1.2in]{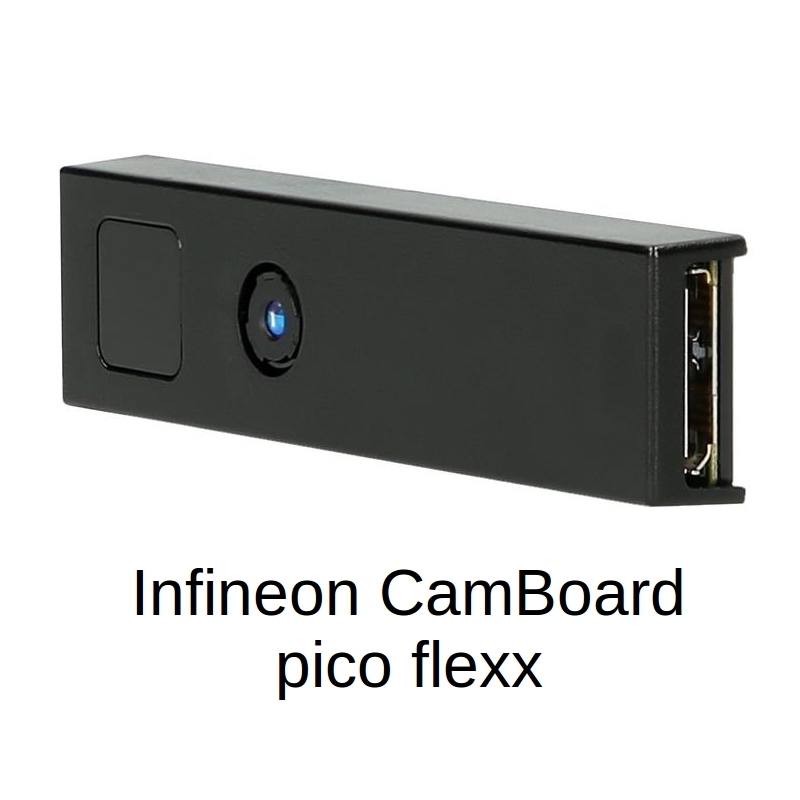}}
    \quad
    \subfigure[Top depth image]{
        \includegraphics[width=1.5in]{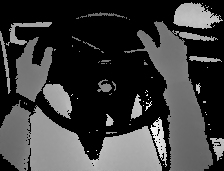}}
    \subfigure[Top infrared image]{
        \includegraphics[width=1.5in]{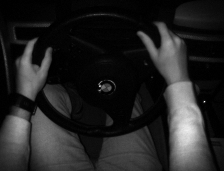}}
    \quad
    \subfigure[Front depth image]{
        \includegraphics[width=1.5in]{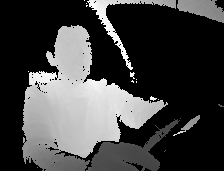}}
    \subfigure[Front infrared image]{
        \includegraphics[width=1.5in]{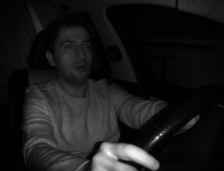}}
	\caption{Environment for data collection. (a) Driving simulator with camera placements. (b) Infineon CamBoard pico flexx camera installed for front and top views. Examples of (c) top depth, (d) top infrared, (e) front depth and (f) front infrared recordings.}
    \label{fig:installation}
\end{figure}

There are several vision-based driver monitoring datasets that are publicly available, but for the task of open set recognition such that normal driving should still be distinguished from unseen anomalous actions, there has been none. In order to fill this research gap, we have recorded the Driver Anomaly Detection (DAD) dataset, which contains the following properties:
\begin{itemize}
	\item The DAD dataset is large enough to train a Deep Neural Network architectures from scratch.
	\item The DAD dataset is multi-modal containing depth and infrared modalities such that system is operable at different lightning conditions.
	\item The DAD dataset is multi-view containing front and top views. These two views are recorded synchronously and complement each other.
	\item The videos are recorded with 45 frame-per-second providing high temporal resolution.
\end{itemize}

We have recorded the DAD dataset using a driving simulator that is shown in Fig. \ref{fig:installation}. The driving simulator contains a real BMW car cockpit, and the subjects are instructed to drive in a computer game that is projected in front of the car. Two Infineon CamBoard pico flexx cameras are placed on top and in front of the driver. The front camera is installed to record the drivers' head, body and visible part of the hands (left hand is mostly obscured by the driving wheel), while top camera is installed to focus on the drivers' hand movements. The dataset is recorded in synchronized depth and infrared modalities with the resolution of \mbox{224 x 171} pixels and frame rate of 45 fps. Example recordings for the two views and two modalities are shown in Fig. \ref{fig:installation}.

\begin{figure}[t!]
	\centering
	\includegraphics[width = 0.48\textwidth]{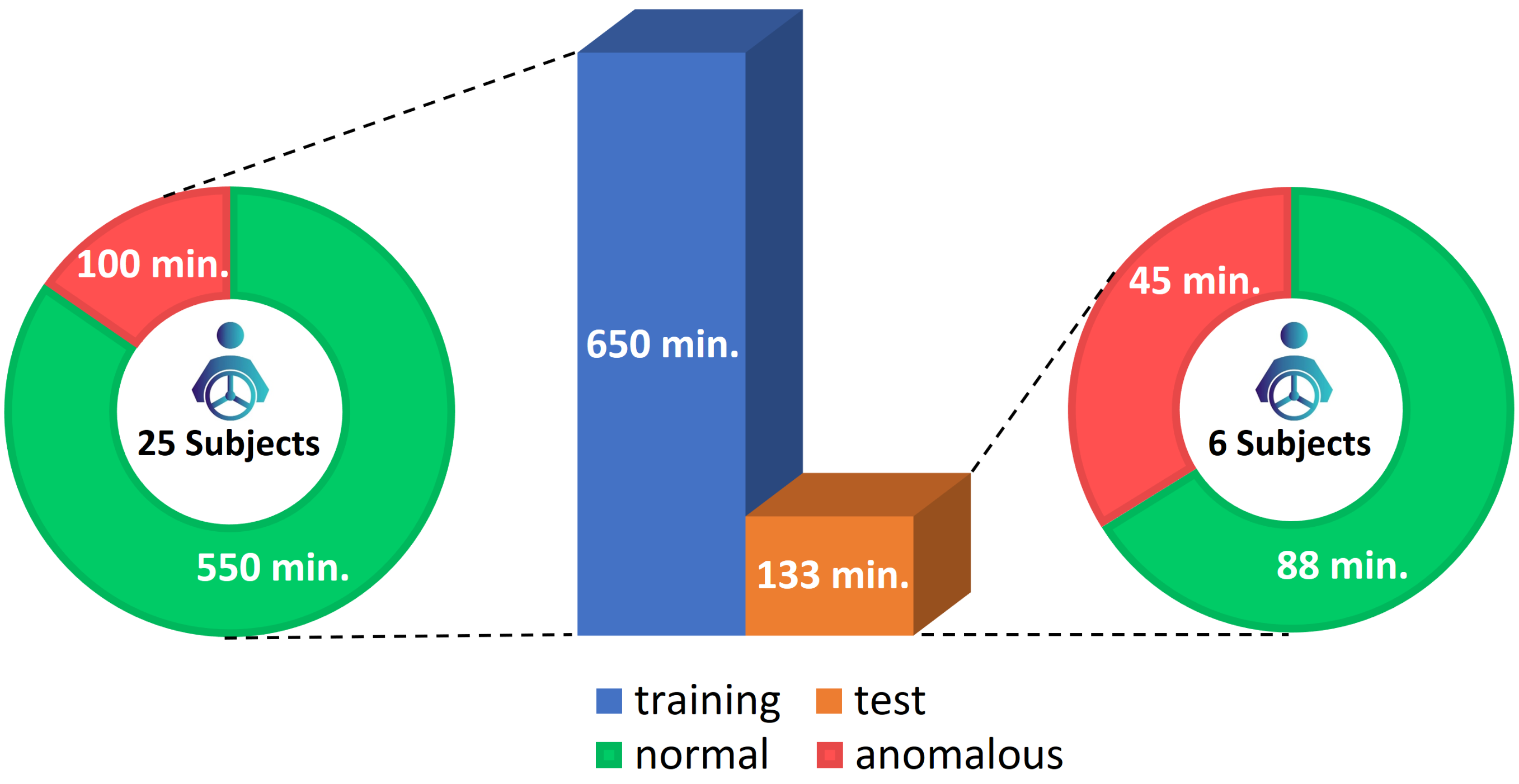}
	\caption{The DAD dataset statistics.}
	\vspace{-0.3cm}
	\label{fig:statistics}
\end{figure}

\begin{table*}[t!]
	\centering
	\resizebox{\textwidth}{!}{
		\begin{tabular}{c|ccc}
			\hline
			\multirow{2}*{\textbf{Anomalous Actions in Training Set}} & \multicolumn{3}{c}{\multirow{2}*{\textbf{Anomalous Actions in Test Set}}}\\ \\
			\hline
			Talking on the phone-left & Talking on the phone-left & \color{red} Adjusting side mirror & \color{red} Wearing glasses \\
			Talking on the phone-right & Talking on the phone-right & \color{red} Adjusting clothes & \color{red} Taking off glasses \\
			Messaging left & Messaging left & \color{red} Adjusting glasses & \color{red} Picking up something\\
			Messaging right & Messaging right & \color{red} Adjusting rear-view mirror & \color{red} Wiping sweat \\
			Talking with passengers & Talking with passengers & \color{red} Adjusting sunroof & \color{red} Touching face/hair \\
			Reaching behind & Reaching behind & \color{red} Wiping nose & \color{red} Sneezing  \\
			Adjusting radio & Adjusting radio & \color{red} Head dropping (dozing off) & \color{red} Coughing \\
			Drinking & Drinking & \color{red} Eating & \color{red} Reading  \\
			\hline
	\end{tabular}}
	\caption{Anomalous actions in the training and test sets. 16 actions in the test set that are not available in the training set are highlighted in red color.}
	\label{tab:classes}
\end{table*}

For the dataset recording, 31 subjects are asked to drive in a computer game performing either \textit{normal driving} or \textit{anomalous driving}. Each subject belongs to either training or test set. The training set contains recordings of 25 subjects and each subject has 6 normal driving and 8 anomalous driving video recordings. Each normal driving video lasts about 3.5 minutes and each anomalous driving video lasts about 30 seconds containing a different distracting action. The list of distracting actions recorded in the training set can be found in Table \ref{tab:classes}. In total, there are around 550 minutes recording for normal driving and 100 minutes recording of anomalous driving in the training set.

The test set contains 6 subjects and each subject has 6 video recordings lasting around 3.5 minutes. Anomalous actions occur randomly during the test video recordings. Most importantly, there are 16 distracting actions in the test set that are not available in the training set, which can be found in Table~\ref{tab:classes}. \textit{Because of these additional distracting actions, the networks need to be trained according to open set recognition task and distinguish normal driving no matter what the distracting action is.} The complete test consists of 88 minutes recording for normal driving and 45 minutes recording of anomalous driving. The test set constitutes the 17\% of the complete DAD dataset, which is around 95 GB. The dataset statistics can be found in Fig. \ref{fig:statistics}.

\section{Methodology}
\label{sec:methodology}

\subsection{Contrastive Learning Framework}

\begin{figure*}[t!]
	\centering
	\includegraphics[width=0.95\textwidth]{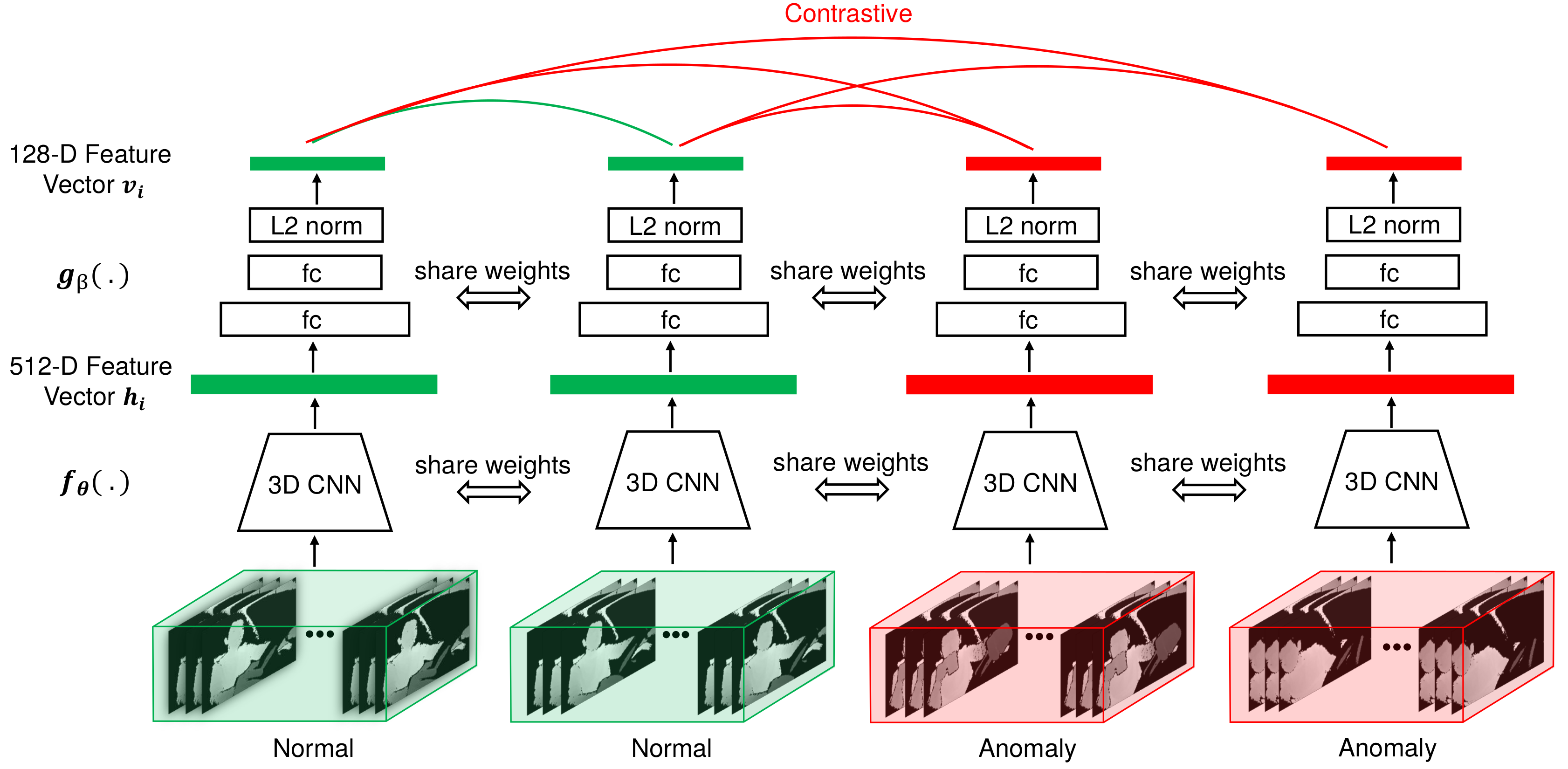}
	\caption{Contrastive learning framework for driver anomaly detection task. A pair of normal driving clips a number of anomaly driving clips (2 in this example) are fed to a base encoder $f_{\theta}$(.) and projection head $g_{\beta}$(.) to extract visual representations of $\boldsymbol h_{\boldsymbol i}$ and $\boldsymbol v_{\boldsymbol i}$, respectively. Once training is completed, projection head is removed, and only the encoder $f_{\theta}$(.) is used for test time recognition.}
	\label{fig:whole_arch}
\end{figure*}

Our motivation is to learn a compact representation for normal driving such that any action deviating from normal driving beyond a threshold can be detected as anomalous action. Accordingly, Inspired by recent progress in contrastive learning algorithms, we try to maximize the similarity between normal driving samples and minimizing the similarity between normal driving and anomalous driving samples in the latent space using a contrastive loss. Fig.~\ref{fig:whole_arch} illustrates the applied framework, which has three major components:

\begin{itemize}
	\item\textbf{Base encoder $\boldsymbol f_{\boldsymbol \theta}$(.)} is used to extract vector representations of input clips. $f_{\theta}$(.) refers to a 3D-CNN architecture with parameters $\theta$. We performed experiments with ResNet-18 and various resource efficient 3D-CNNs to transform input $\boldsymbol x_{\boldsymbol i}$ into $\boldsymbol h_{\boldsymbol i} \in \mathbb{R}^{512}$ via $\boldsymbol h_{\boldsymbol i} = f_\theta(\boldsymbol x_{\boldsymbol i})$.
	\item\textbf{Projection head $\boldsymbol g_{\boldsymbol \beta}$(.)} is used to map $\boldsymbol h_{\boldsymbol i}$ into another latent space $\boldsymbol v_{\boldsymbol i}$. According to findings in \cite{chen2020simple}, it beneficial to define the contrastive loss on $\boldsymbol v_{\boldsymbol i}$ rather than $\boldsymbol h_{\boldsymbol i}$. $\boldsymbol g_{\boldsymbol \beta}$\textbf{(.)} refers to MLP with one hidden layer with ReLU activation and has parameters $\beta$ to achieve transformation of $\boldsymbol v_{\boldsymbol i} = g_{\beta}(\boldsymbol h_{\boldsymbol i}) = W^{(2)}max(0, {W^{(1)}}\boldsymbol h_{\boldsymbol i})$, where $\boldsymbol v_{\boldsymbol i} \in \mathbb{R}^{128}$. After MLP, $\ell$2 normalization is applied to the embedding $\boldsymbol v_{\boldsymbol i}$.
	\item\textbf{Contrastive loss} is used to impose that normalized embeddings from the normal driving class are closer together than embeddings from different anomalous action classes. For this reason, positive pairs in the contrastive loss are always selected from normal driving clips, whereas anomalous driving clips are used only as negative samples. 
\end{itemize}

We divide our normal and anomalous videos into clips for the training. Within a mini-batch, we have $K$ normal driving clips and $M$ anomalous driving clips with index $i \in \{1, ..., K\!+\!M\}$. Final embedding of the $i^{th}$ normal and anomalous driving clips are denoted as $\boldsymbol{v_{ni}}$ and $\boldsymbol{v_{ai}}$, respectively. There are in total $K(K\!-\!1)$ positive pairs and $KM$ negative pairs in every mini-batch. For the supervised contrastive learning approach that we have applied for the task of driver anomaly detection task, the loss takes the following final form:

\begin{equation}
\mathcal{L}_{ij} = - \log \frac{exp(\boldsymbol{v_{ni}}^\mathrm{T}\boldsymbol{v_{nj}}/\tau)}{exp(\boldsymbol{v_{ni}}^\mathrm{T}\boldsymbol{v_{nj}}/\tau)+\sum\limits_{m=1}^{M}exp(\boldsymbol{v_{ni}}^\mathrm{T}\boldsymbol{v_{am}}/\tau)}
\label{equpartial}
\end{equation}
\begin{equation}
\mathcal{L} = \frac{1}{K(K-1)} \sum_{i=1}^{K}\sum_{j=1}^{K}\mathbbm{1}_{j\neq i}\mathcal{L}_{ij}
\label{equtotal}
\end{equation}

\noindent where $\mathbbm{1}$ $\in$ \{0, 1\} is an indicator function that returns 1 if $j \neq i$ and 0 otherwise, and $\tau$ $\in$ (0, $\infty$) is a scalar temperature parameter that can control the concentration level of the distribution \cite{hinton2015distilling}. Typically, $\tau$ is chosen between 0 and 1 to amplify the similarity between samples, that is beneficial for training. The inner product of vectors measures the cosine similarity between encoded feature vectors because they are all $\ell2$ normalized. By optimizing Eq.~(\ref{equtotal}), the encoder is updated to maximize the similarity between the normal driving feature vectors $\boldsymbol{v_{ni}}$ and $\boldsymbol{v_{nj}}$ while minimizing the similarity between the normal driving feature vector $\boldsymbol{v_{ni}}$ and all other anomalous driving feature vectors $\boldsymbol{v_{am}}$ in the same mini-batch. 

\vspace{0.28cm}
\textbf{Noise Contrastive Estimation.} The representation learnt by Eq.~(\ref{equtotal}) can be improved by introducing many more anomaly driving clips (i.e. negative samples). In the extreme case, we can use the complete training samples of the anomalous driving. However, this is too expensive considering the limited memory of the used GPU. Noise Contrastive Estimation \cite{Gutmann2010NoisecontrastiveEA} can be used to approximate the full softmax distribution as in \cite{Gutmann2010NoisecontrastiveEA,wu2018unsupervised}. In our implementation, we have used the $m$ negative samples in our mini-batch and applied ($m$+1)-way softmax classification as also used \cite{tian2019contrastive,he2019momentum,bachman2019learning}. Different from these works, we do not use a memory bank and optimize our framework using only the elements in the mini-batch.

\subsection{Test Time Recognition}
\label{sec:classifier}

The common practice to evaluate learned representations is to train a linear classifier on top of the frozen base network \cite{tian2019contrastive,he2019momentum,bachman2019learning,chen2020simple}. However, this final training is tricky since representations learned by unsupervised and supervised training can be quite different. For example, training of the final linear classification is performed with learning rate of 30, although unsupervised learning is performed with initial learning rate of 0.01. In addition, authors in \cite{wu2018unsupervised} apply $k$-nearest neighbours (kNN) classification for the final evaluation. However, kNN also requires distance calculation with all training clips for each test clip, which is computationally expensive.

For the test time recognition, we propose an evaluation protocol that does not require neither any further training nor complex computations. After the training phase, we throw away the projection head as in \cite{chen2020simple} and use the trained 3D-CNN model to encode every normal driving training clips $\boldsymbol{x_{i}}$, $i \in \{1,...,N\}$ into a set of $\ell2$ normalized 512-dimensional feature representations. Afterwards, normal driving template vector $\boldsymbol{v_{n}}$ can be calculated with:

\begin{equation}
\boldsymbol{v_{n}} = \frac{1}{N} \sum_{i=1}^{N}\phantom{l} \frac{f_{\theta}(\boldsymbol{x_{i}})}{\lVert f_{\theta}(\boldsymbol{x_{i}}) \rVert_2}
\label{equtemplate}
\end{equation}

To classify a test video clip $\boldsymbol{x_i}$, we encode it again into a $\ell2$ normalized 512-dimensional vector and compute the cosine similarity between the encoded clip and $\boldsymbol{v_{n}}$ by:

\begin{equation}
sim_{i} = \boldsymbol{v_{n}}^\mathrm{T}\phantom{l} \frac{f_{\theta}(\boldsymbol{x_{i}})}{\lVert f_{\theta}(\boldsymbol{x_{i}}) \rVert_2}
\label{equsimilarity}
\end{equation}

Finally, any clip whose similarity score below a threshold, $sim_{i}<\gamma$, is classified as anomalous driving. This way, only a simple vector multiplication is performed for test time evaluation. Moreover, similarity score of the test clip $sim_{i}$ gives the severity of the anomalous behavior.

\vspace{0.2cm}
\noindent\textbf{Fusion of Different Views and Modalities.} The DAD dataset contains front and top views; and depth and infrared modalities. We have trained a separate model for each view and modality and fused them later with decision level fusion. As an example, the fused similarity score for top view depth and infrared modalities is calculated with:

\begin{equation}
    sim^{(top)}_{(DIR)} = \frac{sim^{(top)}_{(D)}+sim^{(top)}_{(IR)}}{2}
\end{equation}

It must be noted that each applied view and modality increases the required memory and inference time, which would be critical for autonomous driving applications.

\begin{table*}[!b]
	\centering
	\resizebox{\textwidth}{!}{
		\begin{tabular}{lcccccccccc}
			\specialrule{.15em}{.3em}{.3em}
			\multicolumn{1}{c}{\multirow{3}{*}{\textbf{Model}}} & \multicolumn{1}{c}{\multirow{3}{*}{\textbf{Loss}}} & \multicolumn{9}{c}{\textbf{AUC}} \\ \cmidrule{3-11}
			\multicolumn{1}{c}{} & \multicolumn{1}{c}{} & \multicolumn{3}{c}{\textbf{Top}} & \multicolumn{3}{c}{\textbf{Front}} & \multicolumn{3}{c}{\textbf{Top+Front}} \\ \cmidrule(lr){3-5} \cmidrule(lr){6-8} \cmidrule(lr){9-11} 
			\multicolumn{1}{c}{} & \multicolumn{1}{c}{} & \textbf{Depth} & \textbf{IR} & \textbf{D+IR}  & \textbf{Depth} & \textbf{IR} & \textbf{D+IR} & \textbf{Depth} & \textbf{IR} & \textbf{D+IR}\\
			\specialrule{.15em}{.3em}{.3em}
			ResNet-18 & CE Loss & 0.7982 & 0.8183 & 0.8384 & 0.8416 & 0.8493 & 0.8816 & 0.8783 & 0.8967 & 0.9190 \\
			ResNet-18 & Weighted CE Loss & 0.8047 & 0.8169 & 0.8399 & 0.8921 & \textbf{0.8808} & 0.9044 & 0.9017 & 0.9070 & 0.9275 \\
			ResNet-18 & Contrastive Loss & \textbf{0.9128} & \textbf{0.8804} & \textbf{0.9166} & \textbf{0.8996} & 0.8695 & \textbf{0.9196} & \textbf{0.9609} & \textbf{0.9321} & \textbf{0.9655} \\
			\specialrule{.15em}{.3em}{.3em}
	\end{tabular}}
	\vspace{-0.1cm}
	\caption{Performance Comparison of contrastive loss, CE loss and weighted CE loss for different views and modalities.}
	\vspace{-0.2cm}
	\label{tab:comparison_loss}
\end{table*}

\subsection{Training Details}

We train our models from scratch for 250 epochs using Stochastic Gradient Descent (SGD) with momentum 0.9 and initial learning rate of 0.01. The learning rate is reduced with a factor of 0.1 every 100 epochs. The DAD dataset videos are divided into non-overlapping 32 frames clips.  In every mini-batch, we have 10 normal driving clips and 150 anomalous driving clips. We have set the temperature $\tau = 0.1$. Several data augmentation methods are applied: multi-scale random cropping, salt and pepper noise, random rotation, random horizontal flip (only for top view). We have used 16 frames input clips, which are downsampled from 32 frames and resized to $112 \times 112$ resolution. At test time, the output score of a 16 frames clip is assigned to the middle frame of the clip (i.e. $8^{th}$ frame). For the evaluation metric, we have mainly used area under the cure (AUC) of the ROC curve since it provides calibration-free measure of detection performance.

We have implemented our code in PyTorch, and all the experiments are done using a single Titan XP GPU.

\section{Experiments}
\label{sec:exp}

\noindent\textbf{Baseline Results.} We have used ResNet-18 as base encoder for the baseline results. All the models in the experiments are trained from scratch unless otherwise specified. For every view and modality, a separate model is trained and individual results as well as fusion results are reported in Table~\ref{tab:baseline_result}. The thresholds that are achieving highest classification accuracy are reported in Table~\ref{tab:baseline_result}. However, true positive rate and false positive rates change according to the applied threshold value. Therefore, we have also reported AUC of the ROC curve for baseline evaluation. 

\begin{figure*}[t!]
	\centering
	\includegraphics[width = 0.92\textwidth]{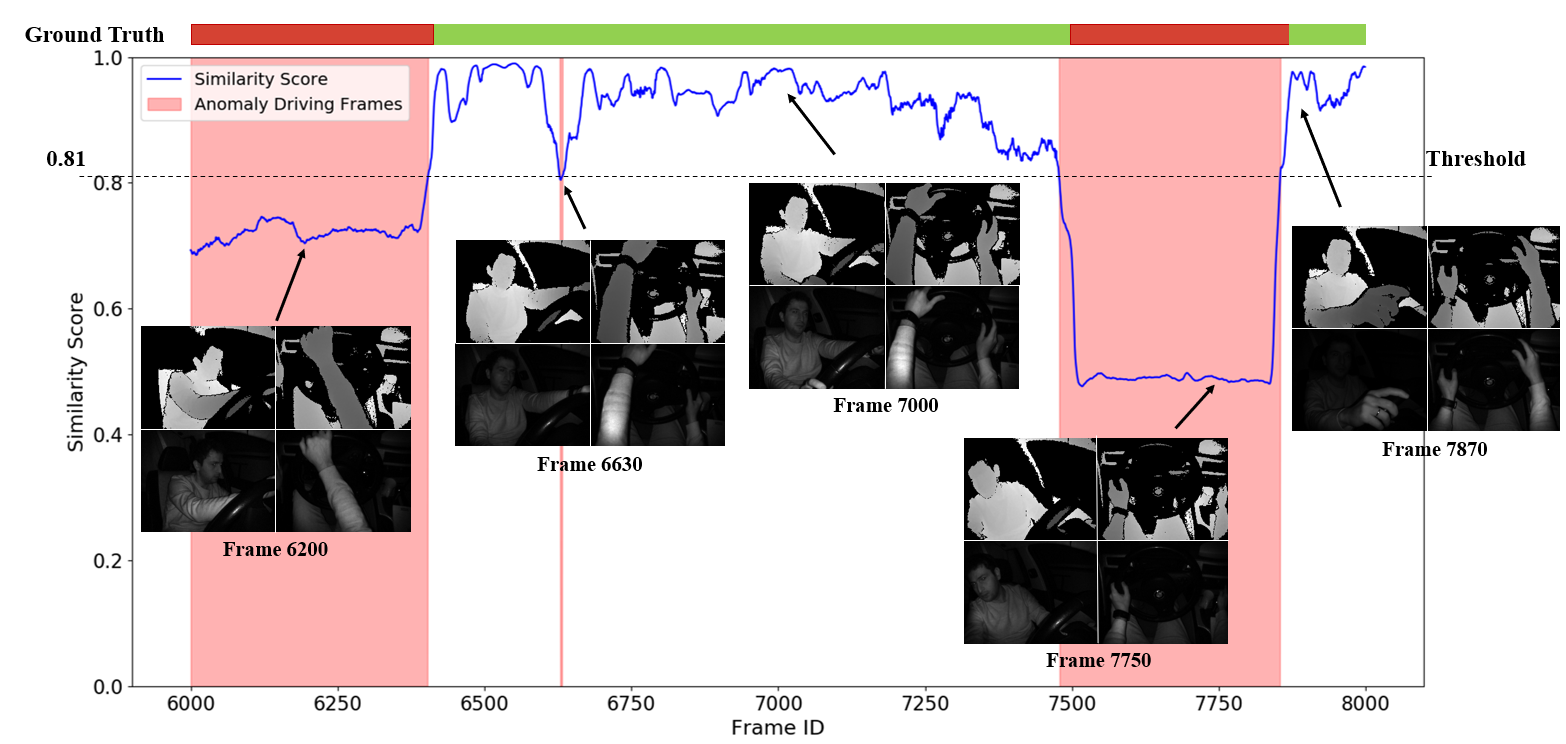}
	\vspace{-0.3cm}
	\caption{Illustration of recognition for a continuous video stream using fusion of both views and modalities. Similarity score refers to cosine similarity between the normal driving template vector and base encoder embedding of input clip. The frames are classified as anomalous driving if the similarity score is blow the preset threshold.}
	\vspace{-0.2cm}
	\label{fig:recognition_illustration}
\end{figure*}

Fusion of different modalities as well as different views always achieves better performance compared to single modalities and views. This shows that different views/modalities in the dataset contains complementary information. Fusion of top/front views and depth/infrared modalities achieves the best performance with 0.9655 AUC. Using this fusion network, the visualization for a continuous video stream is illustrated in Fig.~\ref{fig:recognition_illustration}.

\vspace{0.2cm}
\noindent\textbf{Contrastive Loss or Cross Entropy Loss?} We have compared the performance of contrastive loss and cross entropy (CE) loss. We have trained a ResNet-18 with a final fc layer with CE loss to perform binary classification. However, since the data distribution for normal and anomalous driving is unbalanced in the training set of DAD dataset, we have also experimented with weighted CE loss, where weights are set by inverse class frequency. Comperative results are reported in Table~\ref{tab:comparison_loss}. Our findings are in accordance with \cite{khosla2020supervised}. Except for front view infrared modality, contrastive loss always outperforms CE loss.

\begin{table}[t!]
\centering
\begin{tabular}{lccc}
\specialrule{.15em}{.3em}{.3em}
\multirow{1}*{\textbf{Metric}} & \multirow{1}*{\textbf{Thresholds $\gamma$}} & \multirow{1}*{\textbf{Acc. (\%)}}   & \multirow{1}*{\textbf{AUC}} \\
\specialrule{.15em}{.3em}{.3em}
Top(D)          & 0.89  & 89.13  &0.9128 \\
Top(IR)         & 0.65  & 83.63  &0.8804 \\
Top(DIR)        & 0.76  & 87.75  &0.9166 \\
\specialrule{.1em}{.3em}{.3em}
Front(D)        & 0.75  & 87.21  &0.8996\\
Front(IR)       & 0.82  & 83.68  &0.8695 \\
Front(DIR)      & 0.81  & 88.68  &0.9196 \\
\specialrule{.1em}{.3em}{.3em}
Top+Front(D)    & 0.83  & 91.60  &0.9609 \\
Top+Front(IR)   & 0.80  & 87.06  &0.9311 \\
Top+Front(DIR)  & 0.81  & \textbf{92.34} & \textbf{0.9655} \\
\specialrule{.15em}{.3em}{.3em}
\end{tabular}
\vspace{-0.3cm}
\caption{Results obtained by using a ResNet-18 as base encoder. Thresholds that result in highest classification accuracy are reported.}
\vspace{-0.2cm}
\label{tab:baseline_result}
\end{table}

\begin{table*}[h!]
	\centering
	\resizebox{\textwidth}{!}{
		\begin{tabular}{lccccccccccc}
			\specialrule{.15em}{.3em}{.3em}
			\multicolumn{1}{c}{\multirow{3}{*}{\textbf{Model}}} & \multicolumn{1}{c}{\multirow{3}{*}{\textbf{Params}}} & \multicolumn{1}{l}{\multirow{3}{*}{\textbf{MFLOPS}}} & \multicolumn{9}{c}{\textbf{AUC}} \\ \cmidrule(lr){4-12} 
			\multicolumn{1}{c}{} & \multicolumn{1}{l}{} & \multicolumn{1}{l}{} & \multicolumn{3}{c}{\textbf{Top}} & \multicolumn{3}{c}{\textbf{Front}} & \multicolumn{3}{c}{\textbf{Top+Front}} \\ \cmidrule(lr){4-6} \cmidrule(lr){7-9} \cmidrule(lr){10-12} 
			\multicolumn{1}{c}{} & \multicolumn{1}{l}{} & \multicolumn{1}{l}{} & \multicolumn{1}{l}{\textbf{Depth}} & \multicolumn{1}{c}{\textbf{IR}} & \multicolumn{1}{l}{\textbf{D+IR}} & \multicolumn{1}{l}{\textbf{Depth}} & \multicolumn{1}{c}{\textbf{IR}} & \multicolumn{1}{l}{\textbf{D+IR}} & \multicolumn{1}{l}{\textbf{Depth}} & \multicolumn{1}{c}{\textbf{IR}} & \multicolumn{1}{l}{\textbf{D+IR}} \\
			\specialrule{.15em}{.3em}{.3em}
			MobileNetV1 2.0x  & 13.92M & 499 &0.9125  & 0.8381 & \multicolumn{1}{c}{0.9097} & 0.9018  & 0.8374 & \multicolumn{1}{c}{0.9057} &0.9474  &0.9059  &0.9533  \\
			
			MobileNetV2 1.0x  & 3.01M & 470 & 0.9124 & 0.8531 & \multicolumn{1}{c}{0.9146} & 0.8899 & 0.8355 & \multicolumn{1}{c}{0.8984} & 0.9641 & 0.9154 & 0.9608 \\ 
			
			ShuffleNetV1 2.0x & 4.59M & 413 & 0.8884 & 0.8567 & \multicolumn{1}{c}{0.8926} & 0.8869 & 0.8398 & \multicolumn{1}{c}{0.9000} & 0.9358 & 0.9023 & 0.9480 \\
			
			ShuffleNetV2 2.0x & 6.46M & 383 & 0.8959 & 0.8570 & \multicolumn{1}{c}{0.9066} & 0.9002 & 0.8371 & \multicolumn{1}{c}{0.9054} & 0.9490 & 0.9131 & 0.9531 \\
			\specialrule{.1em}{.3em}{.3em}
			
			ResNet-18 (from scratch) & 32.99M & 6104 & 0.9128 & 0.8804 & \multicolumn{1}{c}{0.9166} & 0.8996 & 0.8695 & \multicolumn{1}{c}{0.9196} & 0.9609 & 0.9311 & 0.9655 \\ 
			
			ResNet-18 (pre-trained) & 32.99M & 6104 & \textbf{0.9200} & \textbf{0.8857} & \multicolumn{1}{c}{\textbf{0.9228}} & \textbf{0.9020} & 0.8666 & \multicolumn{1}{c}{0.9128} & \textbf{0.9646} &0.9227 & 0.9620 \\
			
			ResNet-18 (post-processed) & 32.99M & 6104 & 0.9143 & 0.8827 & \multicolumn{1}{c}{0.9182} & \textbf{0.9020} & \textbf{0.8737} & \multicolumn{1}{c}{\textbf{0.9223}} & 0.9628 & \textbf{0.9335} & \textbf{0.9673}\\
			\specialrule{.15em}{.3em}{.3em}
	\end{tabular}}
	\vspace{-0.3cm}
	\caption{Comparison of different network architectures over AUC, number of parameters and MFLOPS. All architectures takes 16 frames input with $112 \times 112$ spatial resolution.}
	\vspace{-0.2cm}
	\label{tab:different_networks}
\end{table*}

\begin{figure}[t!]
	\centering
	\includegraphics[width=0.45\textwidth]{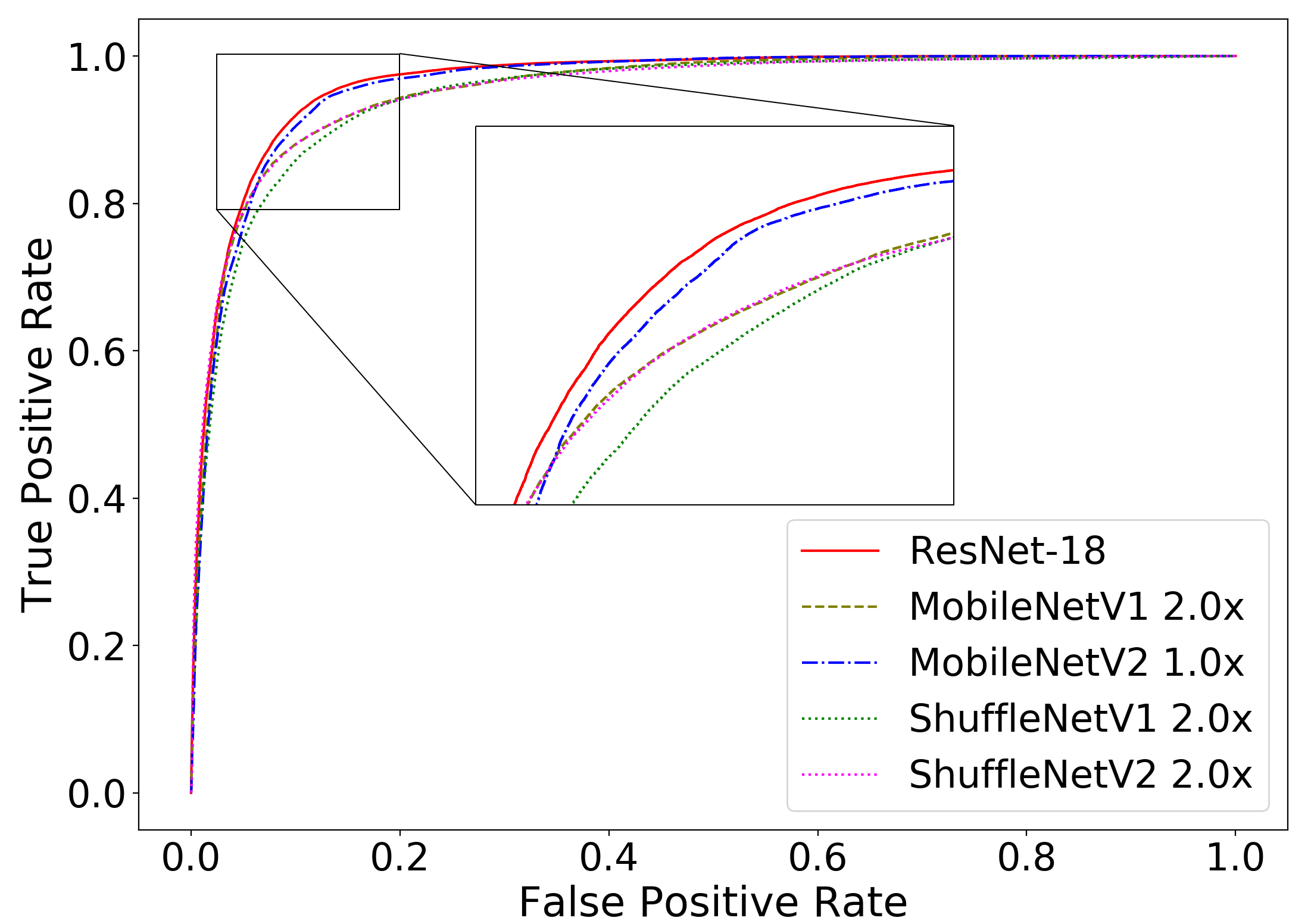}
	\caption{ROC curves using 5 different base encoders. The curves are drawn for the fusion of both views and modalities.}
	\vspace{-0.3cm}
	\label{fig:roc_curve}
\end{figure}

\vspace{0.2cm}
\noindent\textbf{Resource Efficient Base Encoders.} For autonomous applications, it is critical that the deployed systems should be designed considering resource efficiency. Therefore, we have experimented with different resource efficient 3D CNNs \cite{kopuklu2019resource} as base encoder. Comperative results are reported in Table~\ref{tab:different_networks}. Out of all resource efficient 3D CNNs, MobileNetV2 stands out with its performance achieving close to ResNet-18 architecture. More importantly, MobileNetV2 has around 11 times less parameters and requires 13 times less computation compared to ResNet-18. ROC curves for different base encoders are also depicted in Fig.~\ref{fig:roc_curve}, where ResNet-18 and MobileNetV2 again stands out in terms of performance compared to other networks.

\vspace{0.2cm}
\noindent\textbf{With or Without Pre-training?} Transfer learning is a common and effective strategy to improve generalization in small-scale datasets by pretraining network initially with a large-scale dataset \cite{yosinski2014transferable}. Therefore, in order to investigate the effect of pretraining, we have pretrained our \mbox{ResNet-18} base encoder on Kinetics-600 for 100 epochs with contrastive loss similar to our contrastive learning approach described in Section~\ref{sec:methodology}. We have not applied CE loss that is common for training classification tasks since feature representations learnt by CE loss and contrastive loss would be quite different, hence can hinder the transfer learning performance. Before fine-tuning, we have modified the initial convolution layer of the pretrained network to accommodate single channel input by averaging weights of 3 channels. Afterwards, we fine-tune the network using the DAD dataset. Comparative results are reported in Table~\ref{tab:different_networks} that pretrained base encoder does not show apparent advantages compared to base encoder trained from scratch. We infer that our DAD dataset is large enough and the networks that are trained from scratch can already learn all distinctive features without the need of transfer learning.

\vspace{0.2cm}
\noindent\textbf{Post Processing.} It is a common approach to apply post processing in order to prevent fluctuation of detected scores \cite{kopuklu2019real}. For instance, the misclassification between frames 6500 and 6750 in Fig.~\ref{fig:recognition_illustration} can be prevented by such a post processing. Therefore, we have applied a simple low pass filtering (i.e. averaging) on the predicted scores. Instead of making score predictions considering only the current clip, we have applied a running averaging on the $k$-previous scores. We have experimented with different $k$ values and best results are achieved when $k=6$. Comparative results with and without post processing are reported in Table~\ref{tab:different_networks}, where post processing slightly improves the performance.

\begin{table}[b!]
	\centering
	\begin{tabular}{ccc}
		\specialrule{.15em}{.3em}{.3em}
		\begin{tabular}[c]{@{}c@{}}\textbf{Closed-set}\\\textbf{Specificity}\end{tabular} & \begin{tabular}[c]{@{}c@{}}\textbf{Open-set}\\\textbf{Specificity}\end{tabular} & \begin{tabular}[c]{@{}c@{}}\textbf{Average}\\\textbf{Specificity}\end{tabular} \\ 
		\cmidrule(lr){1-3}
		0.8713     & 0.8252     & 0.8565 \\
		\specialrule{.15em}{.3em}{.3em}
	\end{tabular}
	\vspace{-0.3cm}
	\caption{Performance comparison of ResNet-18 on closed-set and open-set anomalies. Fusion of both views and modalities are used, and threshold of 0.81 is applied.}
	\vspace{-0.2cm}
	\label{tab:open-closed-set}
\end{table}

\vspace{0.2cm}
\noindent\textbf{Closed-set and open-set anomalies.} We have compared the performance of proposed architecture over closed-set and open-set anomalies separately in Table \ref{tab:open-closed-set}. According to these results, we can verify that the proposed architecture successfully detects open-set anomalies, although closed-set performance is still better than open-set.

\vspace{0.2cm}
\noindent\textbf{How Training Data Affects the Performance?} The quality and the amount of training data is one of the most important factors on the performance of deep learning applications. Therefore we have investigated the impact of different amounts of training data. First, we have created 5 equal folds each containing training data of 5 subjects. Then, keeping all the anomalous driving in the training set, we have gradually increased the used folds for normal driving data. We have applied the same procedure by switching the normal and anomalous driving subsets. The comparative results are reported in Table~\ref{tab:ratio}, where $\lambda_n$ and $\lambda_a$ refers to the proportion of the used training data for normal driving and anomalous driving subsets, respectively.

The results in Table~\ref{tab:ratio} show that as we increase the amount of normal and anomalous driving videos, achieved performance also increases accordingly. This is natural since we need more normal driving data in order to increase the generalization strength of the learned embeddings. We also need enough anomalous driving data in the training set to draw the boundary of the normal driving embedding and increase the compactness of the learned representation.

\begin{table}[t!]
	\centering
	\begin{tabular}{c|c|ccc}
		\specialrule{.15em}{.3em}{.3em}
		\multicolumn{2}{c}{\textbf{Ratio}} & \multicolumn{3}{c}{\textbf{AUC}} \\ 
		\cmidrule(lr){1-2} \cmidrule(lr){3-5} 
		$\boldsymbol{\lambda_{n}}$ & $\boldsymbol{\lambda_{a}}$ & \phantom{aaa}\textbf{Top}\phantom{aaa} & \phantom{aa}\textbf{Front}\phantom{a} & \textbf{Top+Front} \\ 
		\specialrule{.15em}{.3em}{.3em}
		20\%  & 100\% & 0.7956 & 0.7639 & 0.8513 \\
		40\%  & 100\% & 0.7795 & 0.8111 & 0.8561 \\
		60\%  & 100\% & 0.8599 & 0.8166 & 0.8802 \\
		80\%  & 100\% & 0.8998 & 0.8601 & 0.9382 \\
		100\% & 20\%  & 0.8025 & 0.7873 & 0.8545 \\
		100\% & 40\%  & 0.8103 & 0.8577 & 0.9070 \\
		100\% & 60\%  & 0.8694 & 0.8911 & 0.9335 \\
		100\% & 80\%  & 0.8854 & 0.8921 & 0.9484 \\
		100\% & 100\% & \textbf{0.9128} & \textbf{0.8996} & \textbf{0.9609} \\
		\specialrule{.15em}{.3em}{.3em}
	\end{tabular}
	\caption{Performance comparison using different amount of normal and anomalous driving data in the training. Results are reported for \mbox{ResNet-18} base encoder on depth modality.}
	\label{tab:ratio}
\end{table}

\section{Conclusion}

In this paper, we propose an open set recognition based approach for a driver monitoring application. For this objective, we create and share a video based benchmark dataset, Driver Anomaly Detection (DAD) dataset, which contains unseen anomalous action classes in its test set. Correspondingly, the main task in this dataset is to distinguish normal driving from anomalous driving even some of the anomalous actions have never been seen. We propose a contrastive learning approach in order to generalize the learned embedding of the normal driving video, which can later be used to detect anomalous actions in the test set. 

In our experiments, we have validated that the proposed DAD dataset is large enough to train deep architectures from scratch and has different views and modalities that contain complementary information. Since autonomous applications are limited in terms of hardware, we have also experimented with resource efficient 3D CNN architectures. We specifically note that MobileNetV2 achieves close to \mbox{ResNet-18} performance, but contains 11 times less parameters and requires 13 times less computations than \mbox{ResNet-18}. 

We believe that this work will bring a new perspective to the research on driving monitoring systems. We strongly encourage research community to use open set recognition approaches for detecting drivers' distraction.
\section*{Acknowledgements}
We gratefully acknowledge the support of NVIDIA Corporation with the donation of the Titan Xp GPU, and Infineon Technologies with the donation of Pico Flexx ToF cameras used for this research.

{\small
\bibliographystyle{ieee_fullname}
\bibliography{egbib}

\begin{thebibliography}{10}\itemsep=-1pt

\bibitem{abouelnaga2017real}
Yehya Abouelnaga, Hesham~M Eraqi, and Mohamed~N Moustafa.
\newblock Real-time distracted driver posture classification.
\newblock {\em arXiv preprint arXiv:1706.09498}, 2017.

\bibitem{agtzidis2019360}
Ioannis Agtzidis, Mikhail Startsev, and Michael Dorr.
\newblock 360-degree video gaze behaviour: A ground-truth data set and a
  classification algorithm for eye movements.
\newblock In {\em Proceedings of the 27th ACM International Conference on
  Multimedia}, pages 1007--1015, 2019.

\bibitem{bachman2019learning}
Philip Bachman, R~Devon Hjelm, and William Buchwalter.
\newblock Learning representations by maximizing mutual information across
  views.
\newblock In {\em Advances in Neural Information Processing Systems}, pages
  15535--15545, 2019.

\bibitem{chen2020simple}
Ting Chen, Simon Kornblith, Mohammad Norouzi, and Geoffrey Hinton.
\newblock A simple framework for contrastive learning of visual
  representations.
\newblock {\em arXiv preprint arXiv:2002.05709}, 2020.

\bibitem{das2015performance}
Nikhil Das, Eshed Ohn-Bar, and Mohan~M Trivedi.
\newblock On performance evaluation of driver hand detection algorithms:
  Challenges, dataset, and metrics.
\newblock In {\em 2015 IEEE 18th International Conference on Intelligent
  Transportation Systems}, pages 2953--2958. IEEE, 2015.

\bibitem{diaz2016reduced}
Katerine Diaz-Chito, Aura Hern{\'a}ndez-Sabat{\'e}, and Antonio~M L{\'o}pez.
\newblock A reduced feature set for driver head pose estimation.
\newblock {\em Applied Soft Computing}, 45:98--107, 2016.

\bibitem{dingus2016driver}
Thomas~A Dingus, Feng Guo, Suzie Lee, Jonathan~F Antin, Miguel Perez, Mindy
  Buchanan-King, and Jonathan Hankey.
\newblock Driver crash risk factors and prevalence evaluation using
  naturalistic driving data.
\newblock {\em Proceedings of the National Academy of Sciences},
  113(10):2636--2641, 2016.

\bibitem{fang2019dada}
Jianwu Fang, Dingxin Yan, Jiahuan Qiao, and Jianru Xue.
\newblock Dada: A large-scale benchmark and model for driver attention
  prediction in accidental scenarios.
\newblock {\em arXiv preprint arXiv:1912.12148}, 2019.

\bibitem{Kaggle}
State Farm.
\newblock State farm distracted driver detection.
\newblock https://www.kaggle.com/c/state-farm-distracted-driver-detection,
  accessed: 22-September-2020.

\bibitem{Gutmann2010NoisecontrastiveEA}
Michael Gutmann and Aapo Hyv{\"a}rinen.
\newblock Noise-contrastive estimation: A new estimation principle for
  unnormalized statistical models.
\newblock In {\em AISTATS}, 2010.

\bibitem{hadsell2006dimensionality}
Raia Hadsell, Sumit Chopra, and Yann LeCun.
\newblock Dimensionality reduction by learning an invariant mapping.
\newblock In {\em 2006 IEEE Computer Society Conference on Computer Vision and
  Pattern Recognition (CVPR'06)}, volume~2, pages 1735--1742. IEEE, 2006.

\bibitem{he2019momentum}
Kaiming He, Haoqi Fan, Yuxin Wu, Saining Xie, and Ross Girshick.
\newblock Momentum contrast for unsupervised visual representation learning.
\newblock In {\em Proceedings of the IEEE/CVF Conference on Computer Vision and
  Pattern Recognition}, pages 9729--9738, 2020.

\bibitem{hinton2015distilling}
Geoffrey Hinton, Oriol Vinyals, and Jeff Dean.
\newblock Distilling the knowledge in a neural network.
\newblock {\em arXiv preprint arXiv:1503.02531}, 2015.

\bibitem{howard2017mobilenets}
Andrew~G Howard, Menglong Zhu, Bo Chen, Dmitry Kalenichenko, Weijun Wang,
  Tobias Weyand, Marco Andreetto, and Hartwig Adam.
\newblock Mobilenets: Efficient convolutional neural networks for mobile vision
  applications.
\newblock {\em arXiv preprint arXiv:1704.04861}, 2017.

\bibitem{iandola2016squeezenet}
Forrest~N Iandola, Song Han, Matthew~W Moskewicz, Khalid Ashraf, William~J
  Dally, and Kurt Keutzer.
\newblock Squeezenet: Alexnet-level accuracy with 50x fewer parameters and< 0.5
  mb model size.
\newblock {\em arXiv preprint arXiv:1602.07360}, 2016.

\bibitem{khosla2020supervised}
Prannay Khosla, Piotr Teterwak, Chen Wang, Aaron Sarna, Yonglong Tian, Phillip
  Isola, Aaron Maschinot, Ce Liu, and Dilip Krishnan.
\newblock Supervised contrastive learning.
\newblock {\em arXiv preprint arXiv:2004.11362}, 2020.

\bibitem{kopuklu2019real}
Okan K{\"o}p{\"u}kl{\"u}, Ahmet Gunduz, Neslihan Kose, and Gerhard Rigoll.
\newblock Real-time hand gesture detection and classification using
  convolutional neural networks.
\newblock In {\em 2019 14th IEEE International Conference on Automatic Face \&
  Gesture Recognition (FG 2019)}, pages 1--8. IEEE, 2019.

\bibitem{kopuklu2019resource}
Okan K{\"o}p{\"u}kl{\"u}, Neslihan Kose, Ahmet Gunduz, and Gerhard Rigoll.
\newblock Resource efficient 3d convolutional neural networks.
\newblock {\em arXiv preprint arXiv:1904.02422}, 2019.

\bibitem{kopuklu2020drivermhg}
Okan K{\"o}p{\"u}kl{\"u}, Thomas Ledwon, Yao Rong, Neslihan Kose, and Gerhard
  Rigoll.
\newblock Drivermhg: A multi-modal dataset for dynamic recognition of driver
  micro hand gestures and a real-time recognition framework.
\newblock {\em arXiv preprint arXiv:2003.00951}, 2020.

\bibitem{kose2019real}
Neslihan Kose, Okan Kopuklu, Alexander Unnervik, and Gerhard Rigoll.
\newblock Real-time driver state monitoring using a cnn based spatio-temporal
  approach.
\newblock In {\em 2019 IEEE Intelligent Transportation Systems Conference
  (ITSC)}, pages 3236--3242. IEEE, 2019.

\bibitem{le2017robust}
T~Hoang~Ngan Le, Kha~Gia Quach, Chenchen Zhu, Chi~Nhan Duong, Khoa Luu, and
  Marios Savvides.
\newblock Robust hand detection and classification in vehicles and in the wild.
\newblock In {\em 2017 IEEE Conference on Computer Vision and Pattern
  Recognition Workshops (CVPRW)}, pages 1203--1210. IEEE, 2017.

\bibitem{ma2018shufflenet}
Ningning Ma, Xiangyu Zhang, Hai-Tao Zheng, and Jian Sun.
\newblock Shufflenet v2: Practical guidelines for efficient cnn architecture
  design.
\newblock In {\em Proceedings of the European Conference on Computer Vision
  (ECCV)}, pages 116--131, 2018.

\bibitem{martin2019drive}
Manuel Martin, Alina Roitberg, Monica Haurilet, Matthias Horne, Simon Rei{\ss},
  Michael Voit, and Rainer Stiefelhagen.
\newblock Drive\&act: A multi-modal dataset for fine-grained driver behavior
  recognition in autonomous vehicles.
\newblock In {\em Proceedings of the IEEE international conference on computer
  vision}, pages 2801--2810, 2019.

\bibitem{ohn2013driver}
Eshed Ohn-Bar, Sujitha Martin, and Mohan Trivedi.
\newblock Driver hand activity analysis in naturalistic driving studies:
  challenges, algorithms, and experimental studies.
\newblock {\em Journal of Electronic Imaging}, 22(4):041119, 2013.

\bibitem{ohn2013vehicle}
Eshed Ohn-Bar and Mohan Trivedi.
\newblock In-vehicle hand activity recognition using integration of regions.
\newblock In {\em 2013 IEEE Intelligent Vehicles Symposium (IV)}, pages
  1034--1039. IEEE, 2013.

\bibitem{ortega2020dmd}
Juan~Diego Ortega, Neslihan Kose, Paola Ca{\~n}as, Min-An Chao, Alexander
  Unnervik, Marcos Nieto, Oihana Otaegui, and Luis Salgado.
\newblock Dmd: A large-scale multi-modal driver monitoring dataset for
  attention and alertness analysis.
\newblock {\em arXiv preprint arXiv:2008.12085}, 2020.

\bibitem{palazzi2018predicting}
Andrea Palazzi, Davide Abati, Francesco Solera, Rita Cucchiara, et~al.
\newblock Predicting the driver's focus of attention: the dr (eye) ve project.
\newblock {\em IEEE transactions on pattern analysis and machine intelligence},
  41(7):1720--1733, 2018.

\bibitem{8187362}
N. {Parikh} and S. {Boyd}.
\newblock {\em Proximal Algorithms}.
\newblock 2014.

\bibitem{roth2019dd}
Markus Roth and Dariu~M Gavrila.
\newblock Dd-pose-a large-scale driver head pose benchmark.
\newblock In {\em 2019 IEEE Intelligent Vehicles Symposium (IV)}, pages
  927--934. IEEE, 2019.

\bibitem{sandler2018mobilenetv2}
Mark Sandler, Andrew Howard, Menglong Zhu, Andrey Zhmoginov, and Liang-Chieh
  Chen.
\newblock Mobilenetv2: Inverted residuals and linear bottlenecks.
\newblock In {\em Proceedings of the IEEE Conference on Computer Vision and
  Pattern Recognition}, pages 4510--4520, 2018.

\bibitem{6365193}
W.~J. {Scheirer}, A. {de Rezende Rocha}, A. {Sapkota}, and T.~E. {Boult}.
\newblock Toward open set recognition.
\newblock {\em IEEE Transactions on Pattern Analysis and Machine Intelligence},
  35(7):1757--1772, 2013.

\bibitem{schwarz2017driveahead}
Anke Schwarz, Monica Haurilet, Manuel Martinez, and Rainer Stiefelhagen.
\newblock Driveahead-a large-scale driver head pose dataset.
\newblock In {\em Proceedings of the IEEE Conference on Computer Vision and
  Pattern Recognition Workshops}, pages 1--10, 2017.

\bibitem{tian2019contrastive}
Yonglong Tian, Dilip Krishnan, and Phillip Isola.
\newblock Contrastive multiview coding.
\newblock {\em arXiv preprint arXiv:1906.05849}, 2019.

\bibitem{wu2019fbnet}
Bichen Wu, Xiaoliang Dai, Peizhao Zhang, Yanghan Wang, Fei Sun, Yiming Wu,
  Yuandong Tian, Peter Vajda, Yangqing Jia, and Kurt Keutzer.
\newblock Fbnet: Hardware-aware efficient convnet design via differentiable
  neural architecture search.
\newblock In {\em Proceedings of the IEEE Conference on Computer Vision and
  Pattern Recognition}, pages 10734--10742, 2019.

\bibitem{wu2018unsupervised}
Zhirong Wu, Yuanjun Xiong, Stella~X Yu, and Dahua Lin.
\newblock Unsupervised feature learning via non-parametric instance
  discrimination.
\newblock In {\em Proceedings of the IEEE Conference on Computer Vision and
  Pattern Recognition}, pages 3733--3742, 2018.

\bibitem{yosinski2014transferable}
Jason Yosinski, Jeff Clune, Yoshua Bengio, and Hod Lipson.
\newblock How transferable are features in deep neural networks?
\newblock In {\em Advances in neural information processing systems}, pages
  3320--3328, 2014.

\bibitem{zhang2018shufflenet}
Xiangyu Zhang, Xinyu Zhou, Mengxiao Lin, and Jian Sun.
\newblock Shufflenet: An extremely efficient convolutional neural network for
  mobile devices.
\newblock In {\em Proceedings of the IEEE Conference on Computer Vision and
  Pattern Recognition}, pages 6848--6856, 2018.

\bibitem{zhuang2019local}
Chengxu Zhuang, Alex~Lin Zhai, and Daniel Yamins.
\newblock Local aggregation for unsupervised learning of visual embeddings.
\newblock In {\em Proceedings of the IEEE International Conference on Computer
  Vision}, pages 6002--6012, 2019.

\bibitem{zoph2016neural}
Barret Zoph and Quoc~V Le.
\newblock Neural architecture search with reinforcement learning.
\newblock {\em arXiv preprint arXiv:1611.01578}, 2016.

\bibitem{zoph2018learning}
Barret Zoph, Vijay Vasudevan, Jonathon Shlens, and Quoc~V Le.
\newblock Learning transferable architectures for scalable image recognition.
\newblock In {\em Proceedings of the IEEE conference on computer vision and
  pattern recognition}, pages 8697--8710, 2018.

\end{thebibliography}
}

\end{document}